%% file: main.tex
\documentclass[manuscript]{acmart}

\setcopyright{none}
\renewcommand\footnotetextcopyrightpermission[1]{}
\acmConference[ICPM 2027]{9th International Conference on Process Mining}{February 2027}{}
\acmYear{2027}
\acmISBN{}
\acmDOI{}

\usepackage{booktabs}
\usepackage{multirow}
\usepackage{hyperref}
\usepackage[subtle]{savetrees}
\usepackage{tikz}
\usetikzlibrary{positioning,arrows.meta,fit,backgrounds,calc}

\newcommand{\Log}{\ensuremath{\mathcal{L}}}
\newcommand{\Acts}{\ensuremath{\mathcal{A}}}
\newcommand{\IB}{\ensuremath{\mathit{IB}}}
\newcommand{\OB}{\ensuremath{\mathit{OB}}}

\newcommand{\smark}{\ensuremath{\blacktriangleright}}
\newcommand{\emark}{\ensuremath{\blacksquare}}
\newcommand{\tD}{\ensuremath{\tau_D}}
\newcommand{\tLone}{\ensuremath{\tau_1}}
\newcommand{\tLtwo}{\ensuremath{\tau_2}}
\newcommand{\tPAT}{\ensuremath{\tau_{\mathrm{PAT}}}}
\newcommand{\tDUP}{\ensuremath{\tau_{\mathrm{DUP}}}}

\begin{document}

\title{Concurrency-Aware Process Model Forecasting with Causal Nets}

\author{Yongbo Yu}
\orcid{0009-0004-2964-6611}
\affiliation{%
  \institution{Research Center for Information Systems Engineering (LIRIS), KU Leuven}
  \city{Leuven}
  \country{Belgium}}
\email{yongbo.yu@kuleuven.be}

\author{Jari Peeperkorn}
\orcid{0000-0003-4644-4881}
\affiliation{%
  \institution{Research Center for Information Systems Engineering (LIRIS), KU Leuven}
  \city{Leuven}
  \country{Belgium}}
\email{jari.peeperkorn@kuleuven.be}

\author{Johannes De Smedt}
\orcid{0000-0003-0389-0275}
\affiliation{%
  \institution{Research Center for Information Systems Engineering (LIRIS), KU Leuven}
  \city{Leuven}
  \country{Belgium}}
\email{johannes.desmedt@kuleuven.be}

\author{Jochen De Weerdt}
\orcid{0000-0001-6151-0504}
\affiliation{%
  \institution{Research Center for Information Systems Engineering (LIRIS), KU Leuven}
  \city{Leuven}
  \country{Belgium}}
\email{jochen.deweerdt@kuleuven.be}

\renewcommand{\shortauthors}{Yu et al.}

\begin{abstract}
Process model forecasting (PMF) aims to predict the process model that will characterize a future period, thereby providing a process-level view of how behavior is expected to evolve. Existing PMF methods, however, forecast directly-follows graphs, which cannot explicitly represent concurrency. We extend PMF to causal nets by forecasting time series of relation and binding counts and using these forecasts to reconstruct future process models with AND/XOR semantics. To evaluate the resulting models, we introduce a protocol that accounts for partial traces and constructs the workflow nets required for conformance checking. Experiments on four event logs show that the forecasted models achieve conformance levels close to those of models re-mined from observations in the corresponding future windows. They also outperform static discovery baselines, which retain high precision on the structurally stable log but exhibit substantial precision losses on the other three logs. Filtering infrequent bindings improves most conformance metrics, although it also removes much of the concurrent behavior captured by the models. 
\end{abstract}

%%  CCS concepts  ----------------------------------------

%%  ----------------------------------------


\maketitle

%% ===========================================================================
\section{Introduction}
\label{sec:intro}

A central goal of process mining is to analyze information systems through event logs that record their historical execution.
Process discovery derives models in various forms that summarize behavior observed during a past period, while predictive process monitoring predicts various notions of interest to understand the remainder of an individual running case~\cite{difrancescomarino2018ppm}. 
Process model forecasting (PMF)~\cite{desmedt2023pmf} predicts the process model of a future period as a whole by learning how the components of a process representation evolve over time. It thereby extends traditionally static process discovery with predictive techniques.
Existing PMF methods forecast the temporal evolution of directly-follows graphs (DFGs). The directly-follows relations are extracted from the log as daily count series and forecast to obtain a future DFG~\cite{yu2024multivariate,yu2025benchmark,yu2026tsfm}.
However, a DFG carries little process semantics~\cite{van2019practitioner} and cannot explicitly distinguish concurrent execution from alternative interleavings, because both may induce the same directly-follows relations.
Prior PMF studies therefore identify forecasting a richer process representation as a central open problem~\cite{yu2025benchmark,yu2026tsfm}.

To address this gap, this work proposes to forecast a causal net~\cite{aalst2011cnets}. Each activity of a causal net carries input and output bindings that specify sets of predecessors and successors that may occur together. 
A multi-member binding explicitly encodes a joint obligation, which we treat as modeled concurrency.
We build on Fodina~\cite{fodina}, a widely used discovery algorithm from the Heuristics miner family \cite{weijters2006hm}.
It derives its dependency graph and bindings from explicit counts, allowing us to decompose discovery into daily count series over three relation families and the input and output binding patterns. 
Batch discovery algorithms such as Fodina are normally applied to a sufficiently large log containing complete or largely complete traces, meaning continuing discovery over time has to work with partial cases without disclosing their future. 
We therefore extract the binding series from daily prefix-inclusion sub-logs, which count each binding once and admit no event dated after the day. 
Each series is forecasted with a pretrained foundation model~\cite{ansari2025chronos2}, which prior work found to be competitive for PMF \cite{yu2026tsfm}, using an expanding observation history at successive daily forecast origins.
A weekly causal net is then reconstructed under Fodina's construction rules.

For evaluation, we convert each causal net into a frequency-annotated workflow net. Because cases active in a bounded window are not necessarily complete, we compare the replay of window fragments with the replay of complete traces and adopt complete-trace replay for the main analysis. We report alignment-based conformance~\cite{adriansyah2011alignments} alongside token-based replay~\cite{rozinat2008conformance} and extend entropic relevance~\cite{alkhammash2022er} to frequency-annotated workflow nets, following our earlier adaptation to partial traces~\cite{yu2025benchmark}.
We compare the forecast against three references: one obtained through conventional process discovery on static historical logs, and two constructed using future information for ablation analysis.

Overall, we investigate whether a concurrency-bearing process model can be forecast, and how such a forecast should be measured. This paper contributes:
\begin{enumerate}
  \item A decomposition of causal net discovery into daily count series over three relation families and input/output binding patterns, which makes concurrency forecastable.
  \item A prefix-inclusion windowing scheme and forecasting pipeline that reconstructs concurrency and choice through causal net bindings.
  \item A partial-trace evaluation protocol: replaying the complete trace of each window-active case, silent-transition reduction ($\tau$-reduction), and an entropic relevance-based scoring for frequency-annotated Petri nets.
  \item A four-log study of forecast fidelity against re-mined references, the drift dependence of the static baseline, and the effect of the binding filter on the concurrency that survives.
\end{enumerate}

Section~\ref{sec:background} covers background and related work. Section~\ref{sec:method} presents the decomposition, forecasting and reconstruction of causal nets, and Section~\ref{sec:protocol} the evaluation protocol. Section~\ref{sec:results} reports the experimental setting and the results, Section~\ref{sec:discussion} discusses the findings, their limitations and open problems, and Section~\ref{sec:conclusion} concludes.

%% ===========================================================================
\section{Background and Related Work}
\label{sec:background}

An event log $\Log$ is a multiset of traces over an activity set $\Acts$, where each trace is a finite sequence of timestamped events. Given the events $\Log_{\leq t}$ observed up to time $t$ and a forecast horizon $h$, process model forecasting (PMF) predicts a frequency-annotated model $\widehat{\mathcal{M}}_{t,h}$ of the behavior expected over $[t,t+h]$~\cite{desmedt2023pmf}.
PMF was introduced by treating the edge weights of a DFG as time series, forecasting directly-follows counts over a future window, and reconstructing a graph from the predicted values~\cite{desmedt2023pmf}. Later work varied the forecaster while retaining that target, comparing univariate, multivariate, and learned predictors on the same signals~\cite{yu2024multivariate,yu2025benchmark}, and, more recently, pretrained models that forecast heterogeneous series zero-shot and remain competitive on process relations without per-log training~\cite{ansari2025chronos2,yu2026tsfm}. In all of these studies, the forecast target has remained the same: although the methods aim to predict directly-follows counts more accurately, the forecast model still expresses no more than a DFG can. PMF can help address process drift caused by changes in underlying systems, which may require discovered control-flow models to adapt over time~\cite{pasquadibisceglie2026handling}. Such drift may also require predictive models to be updated over time~\cite{marquez2022updating}. 
We distinguish PMF from adjacent but related tasks: process discovery describes observed behavior; predictive process monitoring forecasts the continuation of an individual case~\cite{difrancescomarino2018ppm}; drift detection identifies changes~\cite{maaradji2015drift}; and streaming discovery updates a model as events arrive~\cite{zelst2018streams}.

A causal net $\mathcal{C}=(\Acts,\IB,\OB)$ assigns each activity $a$ admissible input bindings $\IB(a)$ and output bindings $\OB(a)$~\cite{aalst2011cnets}. Members of a binding participate jointly, whereas distinct bindings represent alternatives: $\OB(a)=\{\{b,c\},\{d\}\}$ allows $a$ to create joint obligations for $b$ and $c$, or an obligation for $d$ alone.
We describe these binding semantics as \emph{AND-like} and \emph{XOR-like}, while assuming no block-structured pairing of splits and joins.
Fodina constructs a causal net from activity, succession, and short-loop counts, first deriving a dependency graph and then retaining observed bindings compatible with it~\cite{fodina}. This explicit count-based interface supports reconstruction from forecast counts. Prior online discovery algorithms have already used sliding-window variants of the Heuristics Miner~\cite{burattin2014control}; however, they do not explicitly account for evolving bindings.

Conformance checking assesses fitness through token replay or alignments~\cite{rozinat2008conformance,adriansyah2011alignments}, and precision penalizes behavior permitted by the model but absent from the log~\cite{munozgama2010precision}. Entropic relevance (ER) measures the bits required to encode a log under a model's trace distribution, thereby incorporating frequency information~\cite{alkhammash2022er}. Standard complete-trace replay assumes initial and final markings, whereas calendar windows can truncate either end of a case. Online conformance handles running cases through decomposition \cite{vanden2014event,burattin2017framework}, prefix-alignments~\cite{zelst2019prefixalign} or imputing missing prefixes~\cite{zaman2021orphan}. These methods motivate the explicit treatment of partial traces when evaluating a model forecast for a bounded window.

%% ===========================================================================
\section{Decomposing, Forecasting and Reconstructing Causal Nets}
\label{sec:method}

\begin{figure}[!t]
  \centering
  \resizebox{\linewidth}{!}{%
  \begin{tikzpicture}[
      font=\small,
      box/.style={draw, rounded corners=2pt, align=center, minimum height=10mm,
                  inner sep=4pt, fill=black!3},
      emph/.style={box, fill=black!10, thick},
      ref/.style={draw, densely dashed, rounded corners=2pt, align=center,
                  inner sep=4pt, fill=white},
      arr/.style={-{Stealth[length=2.5mm]}, thick},
      refarr/.style={-{Stealth[length=2.5mm]}, thick, densely dashed, black!55}
    ]
    \node[box] (log) {Event log\\$\Log$};
    \node[box, above right=7mm and 8mm of log.east, anchor=west]
      (rel) {Relation series\\$A$, $AB$, $ABA$ (Sec.~\ref{sec:series})};
     \node[box, below right=7mm and 8mm of log.east, anchor=west]
      (win) {Binding series $\IB$, $\OB$\\prefix-inclusion sub-logs (Sec.~\ref{sec:prefix})};
    \path let \p1 = ($(win.east)-(win.west)$) in
      node[box, minimum width=\x1, above right=7mm and 8mm of log.east, anchor=west]
      (rel) {Relation series\\$A$, $AB$, $ABA$ (Sec.~\ref{sec:series})};
    \node[emph, minimum height=20mm, anchor=west]
      (chronos) at ($(rel.east)!0.5!(win.east)+(9mm,0)$)
      {Chronos-2\\zero-shot\\(Sec.~\ref{sec:forecast})};
    \node[box, right=7mm of chronos] (agg) {Window totals\\$\widehat c$ over $h$ days};
    \node[emph, right=7mm of agg] (recon)
      {Fodina reconstruction\\graph, repair, bindings, \tPAT{}\\(Sec.~\ref{sec:forecast})};
    \node[box, right=7mm of recon] (cnet)
      {Causal net $\widehat{\mathcal{C}}$\\$\to$ WF-net $\to$ $\tau$-reduced\\(Sec.~\ref{sec:reduction})};
    \node[emph, right=7mm of cnet] (eval)
      {Conformance $\cdot$ ER\\case-complete replay\\(Sec.~\ref{sec:modes}--\ref{sec:measures})};
    \node[ref, below=8mm of recon.south, anchor=north]
      (refs) {Reference models (Sec.~\ref{sec:refmodels}): \texttt{Re-mined} and \texttt{Direct-mined}\\
              on the window's observed events, \texttt{Static} on the training log};
    \draw[arr] (log.east) -- ++(3.5mm,0) |- (rel.west);
    \draw[arr] (log.east) -- ++(3.5mm,0) |- (win.west);
    \draw[arr] (rel.east) -- (rel.east -| chronos.west);
    \draw[arr] (win.east) -- (win.east -| chronos.west);
    \draw[arr] (chronos) -- (agg);
    \draw[arr] (agg) -- (recon);
    \draw[arr] (recon) -- (cnet);
    \draw[arr] (cnet) -- (eval);
    \draw[refarr] (log.south) |- (refs.west);
    \draw[refarr] (refs.east) -| (eval.south);
  \end{tikzpicture}}
  \caption{
  The pipeline extracts daily counts of activities, direct successions, length-two loops, and input and output bindings. Each series is forecast by Chronos-2, and the predicted window totals are used to reconstruct a causal net following Fodina’s construction rules. The net represents concurrency and choice through bindings and is converted to a workflow net for evaluation. The dashed path shows reference models constructed from observed counts or mined directly from event logs.
  }
  \label{fig:pipeline}
\end{figure}
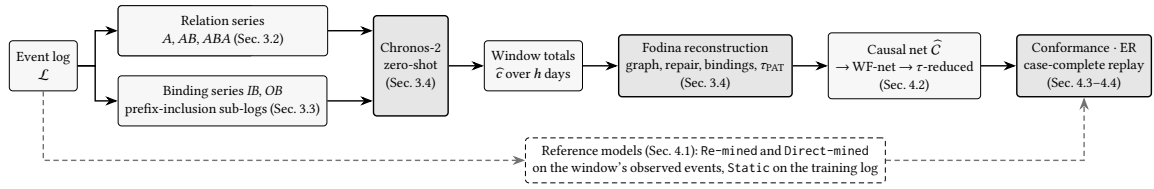

We forecast a causal net by predicting the counts used by its discovery algorithm and rebuilding the net under that algorithm’s construction rules (Figure~\ref{fig:pipeline}). Fodina allows this directly as its dependency graph and its bindings are both derived from explicit occurrence counts~\cite{fodina}. The same strategy may extend to other discovery algorithms with a comparable count-based interface. This section first summarizes the parts of Fodina that define the forecast quantities, then describes how those quantities become daily series, how the binding series are mined without look-ahead, and how the forecasts are turned back into a causal net.

%% ---------------------------------------------------------------------------
\subsection{How Fodina Builds a Causal Net}
\label{sec:fodina}

We write $c(a)$ for the occurrences of activity $a$, $c(a \to b)$ for the direct successions from $a$ to $b$, and $c(a \to b \to a)$ for the length-two loops. Fodina's dependency and loop measures are
\begin{equation}
\label{eq:dep}
\delta(a,b) = \frac{c(a \to b)}{c(a \to b) + c(b \to a) + 1},
\quad
\ell^{1}(a) = \frac{c(a \to a)}{c(a \to a) + 1},
\quad
\ell^{2}(a,b) = \frac{c(a \to b \to a) + c(b \to a \to b)}
                     {c(a \to b \to a) + c(b \to a \to b) + 1},
\end{equation}
with the arc $(a,b)$ admitted when $\delta(a,b) \ge \tD$, the self-loop $(a,a)$ when $\ell^{1}(a) \ge \tLone$, and the pair $\{(a,b),(b,a)\}$ when $\ell^{2}(a,b) \ge \tLtwo$. The additive constant that smooths each measure turns its threshold into an absolute floor. Where $b$ never precedes $a$, admission reduces to $c(a \to b) \ge \tD / (1 - \tD)$, which is one whole occurrence at the default $\tD = 0.5$. This floor has no practical effect on a complete log, but it matters for forecast counts over a seven-day window (Section~\ref{sec:res_modes_cost}). Activity counts fix which tasks the graph carries, and long-distance dependency mining is off in the defaults we use, so the activity, succession and length-two-loop counts are all that a decomposition must supply.

Bindings are mined per firing over that dependency graph.
For each occurrence of $a$, Fodina’s binding miner searches backward for candidate input members and forward for candidate output members among its dependency-graph neighbors. In each direction, it considers only the nearest occurrence of each candidate in trace order and stops searching when another occurrence of $a$ is encountered.
An intervening-neighbor rule rejects a candidate member $b$ when the graph joins $b$ to an event lying between $b$ and $a$ (a successor of $b$ for an input binding, a predecessor for an output binding), since that event accounts for $b$ better than $a$ does. The strictness of this rule rises with the density of the graph, so the granularity at which the graph is built changes which bindings survive (Section~\ref{sec:res_quality}).

The filter \tPAT{} is applied separately to the input and output bindings of each task $a$. For each occurrence of $a$, the mining procedure above collects the accepted candidate members into a binding pattern $B$. Identical non-empty patterns are grouped across occurrences, and $f_a(B)$ counts how many occurrences yield each pattern. Let $\mathcal{B}_a$ denote the distinct patterns collected for the side being filtered, and let $r_a(B)=f_a(B)/c(a)$ be the count of $B$ relative to the occurrence count of $a$. The threshold is the mean ratio $\bar r_a = \frac{1}{|\mathcal{B}_a|}\sum_{B\in\mathcal{B}_a} r_a(B)$ plus a piecewise adjustment in $\tPAT\in[-1,1]$:
\begin{equation}
\label{eq:tpat}
\theta_a = \bar r_a +
\begin{cases}
  \tPAT\,\bar r_a, & \tPAT \le 0, \\[2pt]
  \tPAT\,(1 - \bar r_a), & \tPAT > 0,
\end{cases}
\qquad
B \text{ retained} \iff r_a(B) \ge \theta_a ,
\end{equation}
after which each dependency-graph neighbor that is not included in any retained pattern is re-added as a singleton (a one-member binding). The setting $\tPAT = 0.0$ (Fodina's default) keeps the patterns at or above the mean of their task side, while $\tPAT = -1.0$ removes the threshold to zero and switches the filter off. The re-add step has a consequence that matters later. Discarding a multi-member binding and restoring its members individually replaces one parallel construct with a set of alternatives, so raising \tPAT{} can replace a discarded concurrent binding with singleton alternatives, thereby changing concurrency into choice.

%% ---------------------------------------------------------------------------
\subsection{Decomposition into Daily Count Series}
\label{sec:series}

Let \Log{} be a timestamped event log over activities \Acts{}. Each counted quantity of Section~\ref{sec:fodina} becomes one univariate series sampled per day. For $a \in \Acts$, $A_a(d)$ counts the occurrences of $a$ on day $d$ over all cases. 
For an ordered pair $(a,b)$, $AB_{a\to b}(d)$ counts occurrences of $b$ on day $d$ whose immediately preceding event in the same case is $a$, where $a$ may occur on an earlier day.
For $a \neq b$, the loop series $ABA_{a \to b \to a}(d)$ counts the occurrences of $\langle a,b,a \rangle$ and attributes each to the day of the closing $a$, so that a series entry records what completed on that day.
Bindings need to be extracted separately because they depend on the surrounding trace.
We mine binding patterns per day and record one series per pair $(t,B)$ of an anchor task and a sorted member set, separately for input and for output bindings. Table~\ref{tab:logs} reports the resulting number of series per log.

%% ---------------------------------------------------------------------------
\subsection{Prefix-Inclusion Binding Mining}
\label{sec:prefix}

A daily slice of events does not carry the context that binding mining needs. The earlier events of an active case lie outside it, and its successors have not arrived. Mining completed traces would restore that context but disclose future events, resulting in information leakage. We therefore build the sub-log of day $d$ by prefix inclusion. A case enters it when it has an event on $d$, and contributes the events it has produced up to and including $d$. For a trace $\sigma = \langle (a_1,d_1), \ldots, (a_n,d_n) \rangle$,
\begin{equation}
\label{eq:prefix}
\sigma^{(d)} = \langle a_i \mid d_i \le d \rangle,
\qquad
\sigma \in \Log_d \iff d \in \{d_1, \ldots, d_n\}.
\end{equation}
This sub-log contains exactly the history available to an observer at the end of day $d$. The prefix is not length-bounded, so a long-running case contributes its whole observed history on each day on which the case records an event, a choice that is revisited in Section~\ref{sec:discussion}.
We build a separate dependency graph for each daily sub-log $\Log_d$ and retain all mined binding patterns. We apply the pattern filter only during reconstruction, so patterns are not removed from the time series simply because their relative frequency is low on a particular day. This preserves low-frequency patterns for forecasting and lets us use the same forecasts for both filter settings ($\tPAT = 0.0$ and $-1.0$).

Two recording rules prevent look-ahead and double counting. An input binding is recorded on the day its anchor task fires, matching the attribution of $A_t(d)$. An output binding is recorded on the first day on which a non-empty successor pattern is identified in the observed prefix. The anchor occurrence is then marked as resolved, and later successors do not revise that record. For example, suppose a case records $a$ on day 1, $b$ on day 2 and $c$ on day 3. If the miner identifies ${b}$ as an output binding of $a$ on day 2, that binding is recorded once, while observing $c$ on day 3 does not extend it to ${b,c}$.

%% ---------------------------------------------------------------------------
\subsection{Forecasting and Reconstruction}
\label{sec:forecast}

We independently forecast each series in a rolling-origin zero-shot setting with online test-prefix updates using Chronos-2~\cite{ansari2025chronos2}, selected based on its competitive performance in prior directly-follows forecasting experiments~\cite{yu2026tsfm}.
%We use rolling-origin zero-shot forecasting with online test-prefix updates. 
%Each series is forecast independently with Chronos-2~\cite{ansari2025chronos2}, whose pretrained parameters remain fixed throughout evaluation.
%We fix the forecaster to Chronos-2, motivated by prior evidence on directly-follows forecasting~\cite{yu2026tsfm}, and focus on causal net reconstruction and evaluation rather than a comparison of forecasting algorithms.
We evaluate a closed-world, oracle-vocabulary setting in which the set of possible identifiers is constructed from the full preprocessed log, including the test period. The experiment therefore evaluates count forecasting and reconstruction, not the discovery of previously unseen identifiers. 
Identifiers not yet observed have all-zero histories at the forecast origin.
For a series $q$ with training history $\mathcal{H}_q$ and test observations $t_{q,1}, \ldots, t_{q,T}$, the forecast indexed by $s$ uses the context $[\mathcal{H}_q; t_{q,1}, \ldots, t_{q,s}]$ to predict days $s+1,\ldots,s+h$.
The next forecast is issued one day later, after appending $t_{q,s+1}$, so consecutive target windows overlap on $h-1$ days. 
The point forecast is the predictive median for each target day, and the window total is $\widehat c_s(q) = \sum_{j=1}^{h} \widehat y_{q,s,j}$.
Forecast window totals are not rounded: positive fractional counts remain eligible for reconstruction, while non-positive totals are discarded.

Reconstruction uses the forecast window totals $\widehat{c_s}$, summed over target days $s+1,\ldots,s+h$, to build the dependency graph through Eq.~\eqref{eq:dep}.
The declared boundary markers serve as anchors when present in the reconstruction task set. Otherwise, start and end anchors are selected using count-based estimates of Fodina’s most-started and most-ended activities.
Before constructing bindings, we apply Fodina’s connectivity repair. At each step, it adds the admissible arc with the highest dependency measure to connect an activity that is not reachable from the start or from which the end is not reachable. Repair stops when connectivity is achieved or no admissible arc remains.
An arc added during repair carries the forecast succession count for the corresponding activity pair, even if its dependency score is below the admission threshold. 
Bindings are constructed after repair so that they account for the added arcs. 
Each predicted pattern with a positive total is then intersected with the reconstructed predecessor or successor set of its task, patterns that become identical after the intersection are merged and their totals summed, a task side that is emptied by this intersection is repopulated with singletons, and the filter of Eq.~\eqref{eq:tpat} applies last. 
Our implementation restricts binding members to graph neighbors and checks that every binding node has incoming and outgoing arcs. When repair reports full connectivity, it also checks that every activity in the reconstruction task set is reachable from the selected start anchor and can reach the selected end anchor.

%% ===========================================================================
\section{Partial-Trace-Aware Evaluation Protocol}
\label{sec:protocol}

A forecast causal net has no unique observable ground-truth model and its evaluation must also cope with partial cases.
The protocol defines three reference models, a conversion to (reduced) workflow nets, three treatments of partial traces, and three conformance families.

%% ---------------------------------------------------------------------------
\subsection{Reference Models}
\label{sec:refmodels}

An event log does not contain a unique ground-truth model for a future period,
so we compare the \texttt{Forecast} model with three reference models using the same discovery thresholds.
\texttt{Re-mined} applies the same decomposition and reconstruction pipeline to actual observed counts from the target window.
\texttt{Direct-mined} applies Fodina directly to the window-restricted event log, without our decomposition and reconstruction steps.
Both use inputs only obtainable in a controlled experimental setup and not in practice, and are added for ablation purposes.
\texttt{Static} applies Fodina once to the full training log and uses the resulting model for every test window, which corresponds to static discovery as it is used in practice. 
This corresponds to a realistic setting if where no updates are done after an initial process discovery.
Because \texttt{Forecast} and \texttt{Re-mined} share the same pipeline, including the recording rules and associated bias described in Section~\ref{sec:prefix}, their comparison measures the effect of replacing observed counts with forecast counts.
\texttt{Direct-mined} provides a less pipeline-dependent observational reference by applying Fodina directly to the window’s event slice, but it still depends on Fodina and its configuration and is not ground truth.

%% ---------------------------------------------------------------------------
\subsection{Workflow-Net Conversion and \texorpdfstring{$\tau$}{tau}-Reduction}
\label{sec:reduction}

The selected conformance metrics require Petri net semantics. We convert every causal net into a workflow net using the standard causal net mapping~\cite{aalst2011cnets}. Each activity becomes a visible transition with dedicated input and output places, each dependency becomes a place, and each input or output binding becomes a silent transition connected to the places of its members. Alternative binding transitions encode choice, and one binding with several members encodes synchronization, so the concurrency survives the mapping. The initial and final markings are placed around the declared boundary activities, and the net is annotated with the forecast frequencies (Figure~\ref{fig:netpair}). The mapping is $\tau$-dense: it contains one silent transition per binding and a silent chain through each dependency place, which complicates conformance checking.

Before the primary evaluation, we therefore reduce each net with two local rules, the series fusions of Murata~\cite{murata1989petri} restricted to silent transitions, applied in deterministic order until neither applies. 
The rules preserve the visible language by the usual series-fusion argument, since the removed transition is silent and the removed place has a single producer or a single consumer, but we did not develop this argument into a formal proof for nets with loops and checked it empirically instead. 
We use reduced nets for the primary evaluation to improve tractability. Section~\ref{sec:res_modes_cost} empirically supports this choice by showing that reduction has little effect on alignment-based conformance scores.
Each net is also audited for places from which the final marking is structurally unreachable, by a backward walk from the final marking over places and transitions, and we call a net with such a place unfinishable. Such a net can still complete on some runs, but the behavior that cannot complete distorts fitness and precision, so Section~\ref{sec:res_modes_cost} reports how many nets carried one.

%% ---------------------------------------------------------------------------
\subsection{Partial Traces and Replay Modes}
\label{sec:modes}

An $h$-day window contains complete, prefix-truncated, suffix-truncated and middle-crossing cases, and complete cases are a minority on all four logs (Table~\ref{tab:logs}). Standard replay assumes that a trace starts at the initial marking and ends in the final marking. A missing prefix therefore produces missing tokens, a missing suffix leaves tokens in intermediate places, and a middle-crossing case incurs both. We compare three treatments on the same window-active cases. \emph{window\_raw} replays the events whose timestamps fall inside the window and applies the standard missing- and remaining-token conditions, which is what standard tools report. \emph{window\_no\_remain} is a diagnostic relabeling of \emph{window\_raw} and not a further replay: it uses the same fragments and the same token counts and only drops the remaining-token condition from the perfect-fit classification, so it changes the per-trace label and nothing that enters the fitness value. \emph{case\_complete} retrieves the full trace of every case active in the window and replays it from the initial to the final marking.

Case completion removes the artificial boundary penalties and applies the same trace semantics to all four models, but it widens the temporal scope. Replay then includes behavior outside the target window, and possibly events from the training period, so its scores are comparative measures over window-active cases and not estimates restricted to the behavior inside the forecast window (Section~\ref{sec:limit_future}). Section~\ref{sec:res_modes_cost} compares the three modes and adopts \emph{case\_complete} for the main comparison.

%% ---------------------------------------------------------------------------
\subsection{Measures and Computability}
\label{sec:measures}

We report three metric families: token-based conformance, alignment-based conformance, and the ER-based diagnostic.
Token-based replay reports pooled log fitness, the share of perfectly fitting traces, ETC precision and their harmonic mean $F_1$~\cite{rozinat2008conformance,berti2019tbr,munozgama2010precision}. With missing, consumed, remaining and produced token totals $m$, $c$, $r$ and $p$ summed over the log, pooled fitness is
\begin{equation}
\label{eq:fitness}
\mathit{fit}(L,N)=\frac{1}{2}\left(1-\frac{\sum m}{\sum c}\right)
                 +\frac{1}{2}\left(1-\frac{\sum r}{\sum p}\right).
\end{equation}
Pooling the token counts prevents a long trace from entering the log score as many independent failures. The token perfect-fit share counts the traces replayed with no missing and no remaining token, that is, the traces that the token-replay procedure completed without missing or remaining tokens. ETC precision admits only the prefixes that token replay can fit, so it is computed on a model-conforming subset of the log. We therefore report the coverage of fitted prefixes beside it and interpret precision jointly with fitness. Optimal-alignment fitness and Align-ETC precision provide a second conformance family~\cite{adriansyah2011alignments,adriansyah2012precision}. Alignments map a non-fitting prefix to a minimum-cost model execution, so they do not discard it, but they require a reachable final marking and cost more to compute. The alignment perfect-fit share counts the traces whose optimal alignment contains no log move and no model move on a visible transition, silent moves being free, so a trace that is token-perfect should ordinarily also be alignment-perfect (Section~\ref{sec:res_quality}).

We also report ER adapted to partial traces~\cite{alkhammash2022er,yu2025benchmark}. ER is computed on the window fragments, so its trace treatment differs from case-complete replay.
To apply ER, we construct an automaton from the converted workflow net. Each automaton state is a $\tau$-closure: the set containing a marking and all markings reachable from it by firing only silent transitions.
Where the same visible label leads from a closure to several target closures, the converter retains the target with the highest accumulated activity frequency.
Transition probabilities are normalized activity frequencies, and the binding frequencies on the silent transitions do not enter them. 
The conversion discards alternative target states and binding-frequency information, so the score evaluates a simplified representation of the net.
In particular, two nets with identical activity frequencies and binding sets receive the same score even if they allocate different frequencies to those bindings. We therefore use ER as supporting frequency evidence, rather than as a measure of the correctness of forecast concurrency.

Each metric family is computed by four separately budgeted evaluators: token replay with ETC precision, alignment fitness, Align-ETC precision, and ER. A window–model–evaluator combination defines one evaluation unit, and a computation that exceeds its wall-time or memory budget is recorded as intractable.
Reported means pool the completed units and are accompanied by coverage and they must be read as conditional summaries wherever tractability differs between models. Token fitness is not invariant to the density of silent transitions (Section~\ref{sec:reduction}), so we compare each evaluator family within itself and do not pool scores across the two levels of reduction. Three different quantities are called coverage in what follows. ETC coverage is the share of prefix mass on which ETC precision is computed, the ER fitting ratio is the share of traces the automaton accepts outright, and unit coverage is the number of evaluation units that returned a value within the budget.

%% ===========================================================================
\section{Experimental Results}
\label{sec:results}

We evaluate the proposed approach through three research questions:
\begin{enumerate}
    \item[\textbf{RQ1:}] How closely do forecast-reconstructed causal nets
    match models reconstructed from observed future counts in conformance
    and binding structure?
    \item[\textbf{RQ2:}] How do the models produced by temporal reconstruction
    compare with static discovery and discovery applied directly to each
    target window?
    \item[\textbf{RQ3:}] How does binding filtering affect concurrency and
    conformance, and how do partial-trace treatment and net reduction affect
    conformance scores and evaluation cost?
\end{enumerate}
%% ---------------------------------------------------------------------------
\subsection{Data and Experimental Setting}
\label{sec:setup}

\begin{table*}[!htbp]
  \centering
  \caption{Characteristics of the four preprocessed event logs, their daily relation and binding count series, and the composition of their 7-day evaluation windows. A case is active in a window if it has at least one event in the window, and the four life-cycle shares are computed over active cases and sum to approximately $100\%$ per column (due to rounding).}
  \label{tab:logs}
  \scriptsize
  \setlength{\tabcolsep}{3pt}
  \renewcommand{\arraystretch}{1.0}
  \begin{tabular}{l *{5}{r} *{5}{r} *{6}{r}}
    \toprule
     & \multicolumn{5}{c}{\emph{Log}} & \multicolumn{5}{c}{\emph{Daily counts, by relation family}} & \multicolumn{6}{c}{\emph{Eval.\ windows ($h=7$\,d, 1-d stride)}} \\
    \cmidrule(lr){2-6} \cmidrule(lr){7-11} \cmidrule(lr){12-17}
     & Cases & Events & Act. & Var. & Med.\,(d)
     & $A$ & $AB$ & $ABA$ & \IB{} & \OB{}
     & $N$ & Cases/win & Compl.\,\% & Pref.-tr.\,\% & Suf.-tr.\,\% & Mid.-cr.\,\% \\
    \midrule
    BPI2017      & 40{,}229  & 248{,}236     & 10 & 29  & 13.9 & 10 & 21  & 0  & 22  & 29  & 58  & 1{,}882  & 12.4 & 44.2 & 32.4 & 10.9 \\
    BPI2019      & 197{,}521 & 1{,}298{,}887 & 32 & 741 & 57.1 & 32 & 149 & 11 & 219 & 249 & 55  & 14{,}855 & \phantom{0}6.1 & 47.3 & 19.9 & 26.7 \\
    Sepsis       & 999       & 16{,}009      & 18 & 790 & \phantom{0}5.1 & 18 & 137 & 21 & 252 & 280 & 64  & 17       & 18.0 & 43.4 & 22.5 & 16.2 \\
    Hosp.\ Bill. & 78{,}848  & 439{,}278     & 17 & 301 & 50.1 & 17 & 71  & 8  & 87  & 87  & 139 & 1{,}707  & 23.1 & 35.1 & 12.5 & 29.4 \\
    \bottomrule
  \end{tabular}
\end{table*}

The four logs are previously used for PMF evaluation~\cite{yu2025benchmark,yu2026tsfm,zenodo_pmf_data}. They span a high-volume structured process, a high-dimensional heterogeneous one, a low-volume clinical one and a long-horizon administrative one (Table~\ref{tab:logs}). 
Preprocessing is identical for each log and is applied to the whole log before the split: a relative variant coverage filter at $10^{-4}$, insertion of the boundary markers \smark{} and \emark{} where the raw log lacks them, and a $10\%$ timespan trim at each end. We then split each log chronologically using an $80/20$ split.
The split falls on events, so a case landing on the split contributes its earlier events to the training side and its later events to the test side, and each daily sub-log is built by prefix inclusion so no day sees an event dated after it. Fodina's defaults are kept unchanged ($\tD = \tLone = \tLtwo = 0.5$, $\tDUP = 0.90$) so that the forecast and its references share thresholds, and we report the binding filter at both $\tPAT = 0.0$, Fodina's default, and $\tPAT = -1.0$, which disables it. The horizon is $h = 7$ days with a one-day stride. 
Median duration exceeds seven days in three logs and is 5.1 days in Sepsis, so a weekly window is not necessarily a natural unit of process behavior on any of them. The causal net target needs $3.8$ to $5.2$ times as many series as a directly-follows target, which needs the $AB$ series alone, and $62$--$75\%$ of them are the binding series that carry the concurrency.

Discovering the models is inexpensive. Series extraction, sub-log construction, per-day binding mining, zero-shot inference over all series and reconstruction of all weekly models take $7$--$71$ minutes per log on one NVIDIA H100 GPU, with a peak of $7.2$\,GB of resident memory. 
Evaluation is expensive and runs on CPU. Each evaluation unit ran in a single process with eight BLAS threads under a budget of 20 minutes of wall time and 12 GB of memory beyond the loaded log, on one Intel Xeon W-2245 workstation running up to four units concurrently.

%% ---------------------------------------------------------------------------
\subsection{Replay, Frequency Annotation, and \texorpdfstring{$\tau$}{tau}-Reduction}
\label{sec:res_modes_cost}

On BPI2017, we compared three replay treatments across all four model families. Raw window fragments yielded only $2$--$4\%$ perfectly fitting traces. Ignoring remaining tokens increased this share to about $45\%$. Case-complete replay raised token fitness from $0.85$--$0.86$ to $0.97$--$0.99$, while alignment fitness rose from $0.61$--$0.65$ to $0.97$--$0.99$. These gains are consistent with the high proportion of truncated cases in each weekly window (Table~\ref{tab:logs}). We therefore use case-complete replay in the main analysis, while acknowledging that it includes events outside the forecast window.

\begin{figure}[!htbp]
  \centering
  \includegraphics[width=0.96\linewidth]{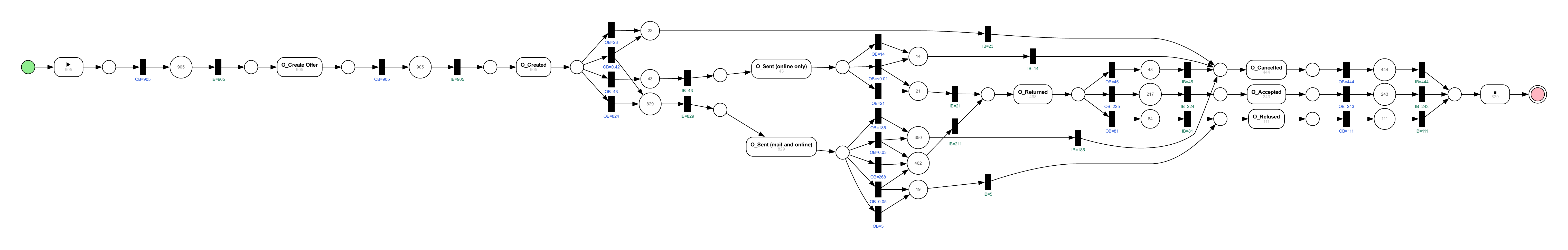}\\[1pt]
  {\scriptsize (a) As constructed: $36$ places, $46$ transitions ($36$ silent), and $96$ arcs.}\par\smallskip
  \includegraphics[width=0.74\linewidth]{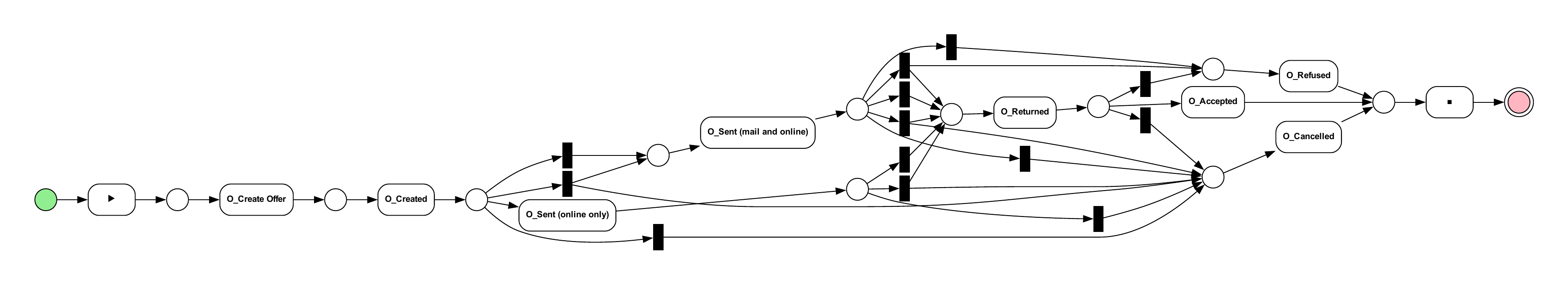}\\[1pt]
  {\scriptsize (b) After $\tau$-reduction: $13$ places, $23$ transitions ($13$ silent), and $50$ arcs.}
  \caption{Forecast workflow net for BPI2017, window~1, with the binding filter disabled ($\tPAT = -1.0$), as constructed (a) and after $\tau$-reduction (b). Circles are places, with the source in green and the sink in pink. Rounded rectangles are visible transitions, black bars are the silent transitions that carry an input (green) or output (blue) binding, and the labels show the forecast frequencies.}
  \label{fig:netpair}
\end{figure}

Figure~\ref{fig:netpair}(a) shows the frequency-annotated Forecast workflow net for the first BPI2017 window with binding filtering disabled. The causal net conversion represents input and output bindings as silent transitions ~\cite{aalst2011cnets}. Labels on visible transitions, dependency places, and silent transitions report forecast window totals for activities, relations, and bindings. Although informative, the conversion and representation are difficult to read and costly to evaluate because it contains many silent transitions. Across the four logs, $5.9\%$ of unreduced evaluation units exceeded the resource budget, with rates of $1.8\%$ at $\tPAT=0.0$ and $10.1\%$ at $\tPAT=-1.0$.
Therefore, we apply a silent-transition reduction. PM4Py's reductions remove fewer silent transitions in this example and can delete a place referenced by the initial or final marking~\cite{pm4py}. Figure~\ref{fig:netpair}(b) shows that our reduction removes many components while retaining the concurrency and choice structure of the example. Across all four logs, it made $47\%$ of previously intractable units computable and reduced evaluation time by $49\%$. Alignment fitness differed by at most $10^{-4}$, while Align-ETC precision and ER were unchanged for $93\%$ and $99.6\%$ of units computed at both levels, respectively. Detailed performance results are available on our GitHub repository.\footnote{\url{https://github.com/YongboYu/pmf-concurrency}}

Token replay provides a fast approximation. However, its ETC precision can be optimistic because non-fitting prefixes are discarded. ETC coverage indicates how much prefix mass supports the reported score. Alignment provides the stricter behavioral view because it evaluates completion to the final marking and aligns non-fitting prefixes. ER instead measures frequency-sensitive coding cost, where lower values are better. All summaries include only successfully computed units sticking to the resource budget. 
Results in Table~\ref{tab:quality} with low coverage provide partial evidence and should be compared with results having similar coverage.

For RQ3, these results show that partial-trace treatment substantially changes conformance scores and that net reduction improves tractability without resolving all evaluation failures. The remaining effect of binding filtering is examined in Section~\ref{sec:res_quality}.

%% ---------------------------------------------------------------------------
\subsection{Conformance Results of the Forecasted models}
\label{sec:res_quality}

\begin{table*}[!htbp]
  \centering
  \caption{Binding structure and case-complete conformance. Conformance uses
           $\tau$-reduced workflow nets and is reported as
           $\mathrm{mean}_{\pm\sigma}$ over tractable units. \emph{AND bind.}
           is the mean per-window percentage of retained input and output
           bindings with multiple members. \texttt{Forecast} and \texttt{Re-mined} use both
           thresholds; \texttt{Direct-mined} and \texttt{Static} use $\tPAT=0.0$. Lower ER
           bits/trace is better. \emph{ETC cov.} is fit-prefix mass.
           \emph{fit. ratio} is the automaton-accepted trace share.
           \emph{no-miss.} ignores remaining tokens. \emph{zero-cost} requires
           completion to the final marking without deviations. \emph{coverage}
           is completed/attempted ER units. $^{\dagger}$ marks fewer than two
           ER values.}
  \label{tab:quality}
  \scriptsize
  \setlength{\tabcolsep}{2.2pt}
  \renewcommand{\arraystretch}{0.95}
  \resizebox{\linewidth}{!}{%
  \begin{tabular}{@{}lll r rrrr rrr rrr@{}}
    \toprule
    & & & \multicolumn{1}{c}{\textbf{Structure}}
      & \multicolumn{4}{c}{\textbf{Token replay}}
      & \multicolumn{3}{c}{\textbf{Alignment}}
      & \multicolumn{3}{c}{\textbf{Entropic relevance}} \\
    \cmidrule(lr){4-4}\cmidrule(lr){5-8}\cmidrule(lr){9-11}
    \cmidrule(lr){12-14}
    \textbf{Log} & \textbf{Model} & $\boldsymbol{\tPAT}$
      & \textbf{AND bind. \%}
      & \textbf{fitness} & \textbf{precision} & \textbf{no-miss. \%}
      & \textbf{ETC cov.}
      & \textbf{fitness} & \textbf{precision} & \textbf{zero-cost \%}
      & \textbf{bits/trace} & \textbf{fit.\ ratio} & \textbf{coverage} \\
    \midrule
    \input{tab-quality-rows-structure}
  \end{tabular}%
  }
\end{table*}

Table~\ref{tab:quality} reports binding structure over windows and conformance over complete traces of cases active in the future window on the reduced nets, at both filter settings for the two reconstruction models and at Fodina's default $\tPAT = 0.0$ for the two mined ones. 

For RQ1, we compare \texttt{Forecast} with \texttt{Re-mined}. At the default binding filter, \texttt{Forecast} achieves precision comparable to \texttt{Re-mined} on BPI2017 and higher precision on the other three logs. Fitness remains broadly similar across all four. 
This indicates that, relative to the available evaluation traces, the forecasted models permit less additional behavior without a substantial loss in measured fitness. One possible explanation is that forecasting suppresses marginal model constructs, effectively regularizing reconstruction, although the present evaluation does not isolate this mechanism.
\texttt{Forecast}'s reconstructed nets also contain fewer tasks and arcs on every log. This pattern is consistent with forecast counts regularizing the reconstruction by suppressing marginal components. 

For RQ2, we compare both reconstruction models with \texttt{Direct-mined} and \texttt{Static}. \texttt{Static} retains high fitness but loses significant precision outside BPI2017. It combines behavior observed throughout the training period and therefore permits many paths that are not relevant to a particular evaluation window. 
On BPI2017, all four models have similarly high fitness and precision with alignment $F_1$ above $0.98$, so these measures reveal little advantage from temporal adaptation. This result is compatible with the log's temporal stability. \texttt{Direct-mined} models retain high fitness but are less precise on several logs, while \texttt{Re-mined} achieves a better fitness–precision balance across all logs, suggesting that the daily-series reconstruction pipeline contributes value independently of forecasting. Because the two families differ in both evidence granularity and model construction, the responsible mechanism cannot be isolated.

Together with Section~\ref{sec:res_modes_cost}, the answer to RQ3 is that filtering, trace treatment and reduction materially affect the reported results. Moving \texttt{Forecast} from $\tPAT=-1.0$ to $0.0$ raises precision and $F_1$ on every log. It also makes ER fully computable on the three logs where the unfiltered nets have poor coverage. This demonstrates that filtering reduces computational complexity, but the filter also removes nearly all multi-member bindings, which are not proved as incorrect, and can replace concurrency with singleton choice. Fitness and precision measure replay behavior, but cannot determine whether the inferred concurrency matches the real process. The filter may remove either false or genuine concurrency that occurs rarely or is weakly represented in the evaluation log.

%% ===========================================================================

\section{Discussion}
\label{sec:discussion}

%% ---------------------------------------------------------------------------
\subsection{Findings}
\label{sec:findings}

RQ1 supports the feasibility of forecasting a causal net through the counts used by its discovery procedure.
Our proposed pipeline makes it possible to forecast causal nets end to end by producing a weekly causal net with explicit joint obligations and alternative bindings from daily count series, and the forecasted net reaches conformance close to a reference re-mined from the events it was predicting. Once discovery is expressed as counts and thresholds, extending PMF to a richer representation reduces to finding the right series to extract and reconstructing under the rules of the discovery algorithm.

RQ2 shows that a model discovered once from the training period can remain competitive, as on BPI2017, but can also become too permissive for an individual window.
In the studied logs, period-specific reconstruction is advantageous where the static (whole-training-period) model loses precision.
A model mined once and reused for each window is competitive on a structurally stable log, but suffers substantial precision losses on the other three. 

RQ3 reveals a tension between retaining concurrent structure and obtaining high conformance scores and tractable evaluation.
How much concurrency a model carries depends on the granularity at which its dependency graph is built and on the binding filter. Whole-period discovery returns models without AND bindings on three logs of four and with more AND structure than per-window mining on the other one with the default pattern filtering. The filter that improves nearly every conformance number is the same filter that removes the concurrency the model was built to carry. 
The reported measures assess the behavior of the complete net but do not directly evaluate whether individual multi-member bindings represent the correct concurrent relations.
Producing a concurrent forecast takes minutes, whereas establishing its quality consumed most of the computational effort of the study. We take these two observations to be the most transferable findings of the study.

%% ---------------------------------------------------------------------------
\subsection{Limitations and Future Work}
\label{sec:limit_future}

This study provides a first proof of concept toward forecasting executable process models rather than a definitive pipeline, while leaving room to further develop the forecasting and evaluation methods required for this research direction.
Applying Fodina to daily sub-logs of incomplete traces changes the evidence used to discover dependencies and bindings.
Fixed smoothing and thresholds may behave differently on sparse daily counts than on a complete log. Mining each day separately can also produce more binding patterns than mining the full log.
Moreover, \tPAT{} responds differently to forecast and observed counts and can replace joint bindings with singleton alternatives, changing concurrency into choice.
Current measures cannot determine whether additional bindings represent meaningful process behavior or distortions introduced by temporal windowing.
Our output-binding recording rule introduces a further limitation. Each activity occurrence is recorded only once, when a non-empty successor pattern is first identified. Successors observed on later days are not added to that record, so a joint binding may be recorded with fewer members than it would have in the completed trace. Comparing these records with bindings mined from completed cases would quantify this effect.
Incomplete traces can also cause Fodina to select an ordinary activity as the end task and suppress its outgoing bindings. Using declared boundary markers during daily mining should therefore be compared with Fodina’s selection of boundaries from the observed traces.
Prefix inclusion uses only events observed by each day, but repeatedly includes the full history of a long-running case on days when it records an event. Older behavior may therefore reduce the influence of recent changes. Future work should evaluate bounded or decay-weighted histories and dependency graphs shared across several days.
Separately, the fixed-vocabulary assumption treats possible relations and bindings as known in advance. Because this vocabulary and the preprocessing use test-period information, the evaluation is not fully prospective, and the effect of these choices remains unquantified. 
Forecast errors can also change model structure and conformance. 
The \texttt{Forecast}/\texttt{Re-mined} comparison assesses their effect under the same reconstruction pipeline, but does not explain how errors in individual series affect the resulting bindings.

The evaluation has additional limitations.
Case-complete replay removes truncation penalties but scores behavior outside the forecast window, making results likely optimistic. Prefix-conditioned replay should initialize from the observed prefix and score only in-window events.
ETC precision can overstate performance at low prefix coverage while ER ignores binding-frequency allocation, does not score predicted counts directly, and becomes unavailable on the most concurrent nets.
To evaluate the semantics of causal nets and DFGs, future studies should compare both targets under identical windows and add binding-level agreement, varied logs, horizons and miners, and synthetic logs with known future models.
More broadly, continuing discovery needs window-adaptive configuration and hierarchical models that reconcile long-run structure with short-run dynamics. 
In this sense, ideas from streaming discovery~\cite{burattin2014control,zelst2018streams} may offer an alternative foundation.

%% ===========================================================================

\section{Conclusion}
\label{sec:conclusion}

In this paper, we showed that a process model carrying concurrency can be forecast end to end. We decomposed causal net discovery into daily count series over three relation families and the input and output binding patterns, extracted the binding series without look-ahead through prefix-inclusion sub-logs, forecast each series zero-shot, and reconstructed a causal net with joint obligations and alternative bindings. Across four event logs, the forecast nets achieved conformance scores on cases active in each future window that were close to those of reference models re-mined from events in that corresponding (future) window. A model mined once from the training period remained competitive on the structurally stable log but suffered substantial precision losses on the other three logs.

Two findings extend beyond this particular pipeline. Producing a concurrent forecast is relatively inexpensive, whereas establishing its quality is not, which is why this paper contributes an evaluation protocol alongside the forecasting pipeline.
The binding filter improved nearly every conformance metric, but also removed much of the concurrency that the models were intended to capture. Moreover, the evaluated measures assess the behavior of the complete net but do not directly determine whether its individual bindings correctly represent concurrency.
Forecasting this richer target is therefore computationally feasible, although its evaluation remains costly. Determining whether the additional concurrent structure is correct requires new binding-aware evaluation methods and constitutes an important next step for concurrency-aware process model forecasting.

\begin{acks}
We acknowledge the use of Generative AI to assist with coding and editing. This work was supported in part by the Research Foundation Flanders (FWO) under Project 1294325N as well as grant number G039923N, and Internal Funds KU Leuven under grant number C14/23/031.
\end{acks}

\bibliographystyle{ACM-Reference-Format}
\bibliography{references}

\end{document}

%% file: tab-quality-rows-structure.tex
    \multirow{6}{*}{BPI2017}
      & \multirow{2}{*}{Forecast} & $-1.0$       & $10.9_{\pm0.6\phantom{0}}$ & $0.915_{\pm0.015}$ & $0.880_{\pm0.014}$ & $\phantom{0}99.6_{\pm0.2\phantom{0}}$ & $100.0_{\pm0.0\phantom{0}}$ & $0.979_{\pm0.024}$ & $0.841_{\pm0.017}$ & $\phantom{0}87.8_{\pm13.2}$ & $\phantom{0}0.663_{\pm0.07\phantom{0}}$ & $0.9996_{\pm0.000}$ & $58/58$ \\
      &                          & $0.0$        & $\phantom{0}0.0_{\pm0.0\phantom{0}}$ & $0.985_{\pm0.018}$ & $0.999_{\pm0.007}$ & $\phantom{0}99.6_{\pm0.2\phantom{0}}$ & $100.0_{\pm0.0\phantom{0}}$ & $0.979_{\pm0.024}$ & $0.999_{\pm0.007}$ & $\phantom{0}87.8_{\pm13.2}$ & $\phantom{0}0.663_{\pm0.07\phantom{0}}$ & $0.9996_{\pm0.000}$ & $58/58$ \\
      & \multirow{2}{*}{Re-mined} & $-1.0$       & $\phantom{0}3.1_{\pm1.7\phantom{0}}$ & $0.937_{\pm0.019}$ & $0.897_{\pm0.025}$ & $\phantom{0}99.9_{\pm0.1\phantom{0}}$ & $100.0_{\pm0.0\phantom{0}}$ & $0.988_{\pm0.014}$ & $0.893_{\pm0.030}$ & $\phantom{0}88.2_{\pm13.2}$ & $\phantom{0}0.691_{\pm0.07\phantom{0}}$ & $1.0000_{\pm0.000}$ & $58/58$ \\
      &                          & $0.0$        & $\phantom{0}0.0_{\pm0.0\phantom{0}}$ & $0.985_{\pm0.017}$ & $0.999_{\pm0.001}$ & $\phantom{0}99.9_{\pm0.1\phantom{0}}$ & $100.0_{\pm0.0\phantom{0}}$ & $0.988_{\pm0.014}$ & $0.999_{\pm0.001}$ & $\phantom{0}88.2_{\pm13.2}$ & $\phantom{0}0.691_{\pm0.07\phantom{0}}$ & $1.0000_{\pm0.000}$ & $58/58$ \\
      \cmidrule(l){2-14}
      & Direct-mined             & $0.0$        & $\phantom{0}0.0_{\pm0.0\phantom{0}}$ & $0.972_{\pm0.035}$ & $0.997_{\pm0.006}$ & $\phantom{0}88.5_{\pm29.9}$ & $\phantom{0}96.7_{\pm8.7\phantom{0}}$ & $0.971_{\pm0.045}$ & $0.997_{\pm0.006}$ & $\phantom{0}76.9_{\pm29.6}$ & $\phantom{0}0.798_{\pm0.32\phantom{0}}$ & $0.9955_{\pm0.012}$ & $58/58$ \\
      & Static                   & $0.0$        & $\phantom{0}0.0_{\pm0.0\phantom{0}}$ & $0.985_{\pm0.018}$ & $0.985_{\pm0.031}$ & $\phantom{0}99.7_{\pm0.2\phantom{0}}$ & $100.0_{\pm0.0\phantom{0}}$ & $0.978_{\pm0.025}$ & $0.985_{\pm0.031}$ & $\phantom{0}88.0_{\pm13.3}$ & $\phantom{0}0.695_{\pm0.07\phantom{0}}$ & $1.0000_{\pm0.000}$ & $58/58$ \\
    \midrule
    \multirow{6}{*}{BPI2019}
      & \multirow{2}{*}{Forecast} & $-1.0$       & $36.2_{\pm1.1\phantom{0}}$ & $0.834_{\pm0.029}$ & $0.754_{\pm0.037}$ & $\phantom{0}32.8_{\pm14.6}$ & $\phantom{0}79.0_{\pm6.9\phantom{0}}$ & $0.944_{\pm0.017}$ & $0.573_{\pm0.021}$ & $\phantom{0}56.5_{\pm10.5}$ & $^{\dagger}$ & $^{\dagger}$ & $\phantom{0}0/55$ \\
      &                          & $0.0$        & $\phantom{0}1.4_{\pm0.7\phantom{0}}$ & $0.939_{\pm0.026}$ & $0.881_{\pm0.031}$ & $\phantom{0}75.6_{\pm14.4}$ & $\phantom{0}90.6_{\pm5.6\phantom{0}}$ & $0.926_{\pm0.026}$ & $0.815_{\pm0.052}$ & $\phantom{0}45.8_{\pm15.1}$ & $\phantom{0}1.894_{\pm0.95\phantom{0}}$ & $0.9525_{\pm0.039}$ & $55/55$ \\
      & \multirow{2}{*}{Re-mined} & $-1.0$       & $13.5_{\pm2.5\phantom{0}}$ & $0.839_{\pm0.040}$ & $0.774_{\pm0.064}$ & $\phantom{0}33.2_{\pm21.9}$ & $\phantom{0}71.3_{\pm12.7}$ & $0.946_{\pm0.017}$ & $0.615_{\pm0.050}$ & $\phantom{0}56.9_{\pm11.3}$ & $^{\dagger}$ & $^{\dagger}$ & $\phantom{0}0/55$ \\
      &                          & $0.0$        & $\phantom{0}2.2_{\pm0.7\phantom{0}}$ & $0.927_{\pm0.023}$ & $0.737_{\pm0.049}$ & $\phantom{0}68.1_{\pm25.0}$ & $\phantom{0}90.1_{\pm6.7\phantom{0}}$ & $0.939_{\pm0.018}$ & $0.661_{\pm0.058}$ & $\phantom{0}51.6_{\pm10.5}$ & $\phantom{0}1.686_{\pm1.14\phantom{0}}$ & $0.9615_{\pm0.046}$ & $14/55$ \\
      \cmidrule(l){2-14}
      & Direct-mined             & $0.0$        & $\phantom{0}1.6_{\pm2.0\phantom{0}}$ & $0.911_{\pm0.025}$ & $0.702_{\pm0.079}$ & $\phantom{0}48.5_{\pm20.6}$ & $\phantom{0}85.0_{\pm13.2}$ & $0.853_{\pm0.088}$ & $0.685_{\pm0.081}$ & $\phantom{0}18.2_{\pm16.6}$ & $\phantom{0}6.149_{\pm1.51\phantom{0}}$ & $0.6984_{\pm0.121}$ & $39/55$ \\
      & Static                   & $0.0$        & $\phantom{0}3.9_{\pm0.0\phantom{0}}$ & $0.906_{\pm0.007}$ & $0.567_{\pm0.026}$ & $\phantom{0}39.5_{\pm10.4}$ & $\phantom{0}85.0_{\pm2.5\phantom{0}}$ & $0.939_{\pm0.017}$ & $0.487_{\pm0.020}$ & $\phantom{0}51.8_{\pm10.5}$ & $^{\dagger}$ & $^{\dagger}$ & $\phantom{0}0/55$ \\
    \midrule
    \multirow{6}{*}{Sepsis}
      & \multirow{2}{*}{Forecast} & $-1.0$       & $17.7_{\pm7.4\phantom{0}}$ & $0.813_{\pm0.058}$ & $0.796_{\pm0.282}$ & $\phantom{00}0.5_{\pm1.7\phantom{0}}$ & $\phantom{0}28.0_{\pm18.3}$ & $0.702_{\pm0.195}$ & $0.749_{\pm0.272}$ & $\phantom{00}0.3_{\pm1.2\phantom{0}}$ & $11.566_{\pm14.08}$ & $0.2014_{\pm0.361}$ & $\phantom{0}8/64$ \\
      &                          & $0.0$        & $\phantom{0}0.1_{\pm0.6\phantom{0}}$ & $0.827_{\pm0.055}$ & $0.820_{\pm0.277}$ & $\phantom{00}0.0_{\pm0.0\phantom{0}}$ & $\phantom{0}27.3_{\pm18.0}$ & $0.677_{\pm0.179}$ & $0.847_{\pm0.259}$ & $\phantom{00}0.0_{\pm0.0\phantom{0}}$ & $24.840_{\pm11.27}$ & $0.3082_{\pm0.226}$ & $64/64$ \\
      & \multirow{2}{*}{Re-mined} & $-1.0$       & $\phantom{0}5.6_{\pm4.5\phantom{0}}$ & $0.825_{\pm0.097}$ & $0.515_{\pm0.255}$ & $\phantom{0}12.9_{\pm26.1}$ & $\phantom{0}37.5_{\pm20.9}$ & $0.645_{\pm0.253}$ & $0.420_{\pm0.270}$ & $\phantom{00}6.3_{\pm5.6\phantom{0}}$ & $13.147_{\pm9.57\phantom{0}}$ & $0.5203_{\pm0.361}$ & $25/64$ \\
      &                          & $0.0$        & $\phantom{0}1.5_{\pm1.7\phantom{0}}$ & $0.834_{\pm0.099}$ & $0.523_{\pm0.253}$ & $\phantom{0}13.7_{\pm26.1}$ & $\phantom{0}37.8_{\pm20.5}$ & $0.631_{\pm0.233}$ & $0.498_{\pm0.276}$ & $\phantom{00}4.9_{\pm4.5\phantom{0}}$ & $20.062_{\pm10.70}$ & $0.4678_{\pm0.290}$ & $46/64$ \\
      \cmidrule(l){2-14}
      & Direct-mined             & $0.0$        & $\phantom{0}1.7_{\pm2.4\phantom{0}}$ & $0.930_{\pm0.062}$ & $0.364_{\pm0.186}$ & $\phantom{0}22.6_{\pm25.2}$ & $\phantom{0}43.7_{\pm21.6}$ & $0.745_{\pm0.237}$ & $0.334_{\pm0.195}$ & $\phantom{0}12.2_{\pm11.2}$ & $16.501_{\pm11.02}$ & $0.6831_{\pm0.226}$ & $46/64$ \\
      & Static                   & $0.0$        & $\phantom{0}0.0_{\pm0.0\phantom{0}}$ & $0.996_{\pm0.002}$ & $0.116_{\pm0.013}$ & $\phantom{0}88.0_{\pm8.0\phantom{0}}$ & $\phantom{0}93.4_{\pm5.1\phantom{0}}$ & $0.993_{\pm0.004}$ & $0.116_{\pm0.013}$ & $\phantom{0}85.2_{\pm7.2\phantom{0}}$ & $13.658_{\pm7.10\phantom{0}}$ & $0.9754_{\pm0.033}$ & $64/64$ \\
    \midrule
    \multirow{6}{*}{Hosp.\ Bill.}
      & \multirow{2}{*}{Forecast} & $-1.0$       & $\phantom{0}9.6_{\pm1.1\phantom{0}}$ & $0.873_{\pm0.018}$ & $0.942_{\pm0.029}$ & $\phantom{0}53.3_{\pm22.8}$ & $\phantom{0}96.4_{\pm0.9\phantom{0}}$ & $0.932_{\pm0.024}$ & $0.871_{\pm0.053}$ & $\phantom{0}56.9_{\pm11.2}$ & $^{\dagger}$ & $^{\dagger}$ & $\phantom{00}1/139$ \\
      &                          & $0.0$        & $\phantom{0}0.0_{\pm0.0\phantom{0}}$ & $0.943_{\pm0.021}$ & $0.977_{\pm0.022}$ & $\phantom{0}93.2_{\pm1.8\phantom{0}}$ & $\phantom{0}96.4_{\pm0.9\phantom{0}}$ & $0.932_{\pm0.024}$ & $0.977_{\pm0.023}$ & $\phantom{0}56.9_{\pm11.2}$ & $\phantom{0}0.577_{\pm0.10\phantom{0}}$ & $0.9895_{\pm0.003}$ & $139/139$ \\
      & \multirow{2}{*}{Re-mined} & $-1.0$       & $\phantom{0}1.0_{\pm1.2\phantom{0}}$ & $0.940_{\pm0.024}$ & $0.884_{\pm0.048}$ & $\phantom{0}86.6_{\pm19.9}$ & $\phantom{0}97.3_{\pm0.8\phantom{0}}$ & $0.949_{\pm0.017}$ & $0.881_{\pm0.050}$ & $\phantom{0}59.3_{\pm11.5}$ & $\phantom{0}0.492_{\pm0.06\phantom{0}}$ & $0.9966_{\pm0.001}$ & $109/139$ \\
      &                          & $0.0$        & $\phantom{0}0.1_{\pm0.4\phantom{0}}$ & $0.948_{\pm0.019}$ & $0.885_{\pm0.047}$ & $\phantom{0}91.1_{\pm16.7}$ & $\phantom{0}97.3_{\pm0.8\phantom{0}}$ & $0.947_{\pm0.016}$ & $0.885_{\pm0.047}$ & $\phantom{0}58.8_{\pm10.8}$ & $\phantom{0}0.507_{\pm0.07\phantom{0}}$ & $0.9962_{\pm0.002}$ & $139/139$ \\
      \cmidrule(l){2-14}
      & Direct-mined             & $0.0$        & $\phantom{0}0.1_{\pm0.4\phantom{0}}$ & $0.956_{\pm0.018}$ & $0.859_{\pm0.081}$ & $\phantom{0}97.5_{\pm1.2\phantom{0}}$ & $\phantom{0}98.5_{\pm0.8\phantom{0}}$ & $0.950_{\pm0.017}$ & $0.859_{\pm0.081}$ & $\phantom{0}59.7_{\pm12.0}$ & $\phantom{0}0.473_{\pm0.22\phantom{0}}$ & $0.9991_{\pm0.002}$ & $124/139$ \\
      & Static                   & $0.0$        & $\phantom{0}0.0_{\pm0.0\phantom{0}}$ & $0.965_{\pm0.014}$ & $0.352_{\pm0.020}$ & $100.0_{\pm0.0\phantom{0}}$ & $100.0_{\pm0.0\phantom{0}}$ & $0.954_{\pm0.018}$ & $0.352_{\pm0.021}$ & $\phantom{0}61.2_{\pm12.8}$ & $\phantom{0}1.521_{\pm0.12\phantom{0}}$ & $0.9979_{\pm0.002}$ & $139/139$ \\
    \bottomrule